\documentclass[conference]{IEEEtran}
\IEEEoverridecommandlockouts
\usepackage{cite}
\usepackage{amsmath,amssymb,amsfonts}
\usepackage{algorithmic}
\usepackage{graphicx}
\usepackage{textcomp}
\usepackage{xcolor}
\usepackage{caption} 
\newcommand{\cev}[1]{\reflectbox{\ensuremath{\vec{\reflectbox{\ensuremath{#1}}}}}}
\usepackage[export]{adjustbox}
\def\BibTeX{{\rm B\kern-.05em{\sc i\kern-.025em b}\kern-.08em
    T\kern-.1667em\lower.7ex\hbox{E}\kern-.125emX}}
\begin{document}

\title{Supervised Transfer Learning for Product Information Question Answering}

\author{
    \IEEEauthorblockN{Tuan Manh Lai\IEEEauthorrefmark{1}, Trung Bui\IEEEauthorrefmark{2}, Nedim Lipka\IEEEauthorrefmark{2}, Sheng Li\IEEEauthorrefmark{3}}
    \IEEEauthorblockA{\IEEEauthorrefmark{1}Purdue University, USA; \IEEEauthorrefmark{2}Adobe Research, USA; \IEEEauthorrefmark{3}University of Georgia, USA\\
    Email: \IEEEauthorrefmark{1}lai123@purdue.edu,
\IEEEauthorrefmark{2}\{bui,lipka\}@adobe.com,
\IEEEauthorrefmark{3}sheng.li@uga.edu}
}

\maketitle

\begin{abstract}
Popular e-commerce websites such as Amazon offer community question answering systems for users to pose product-related questions and experienced customers may provide answers voluntarily. In this paper, we show that the large volume of existing community question answering data can be beneficial when building a system for answering questions related to product facts and specifications. Our experimental results demonstrate that the performance of a model for answering questions related to products listed in the Home Depot website can be improved by a large margin via a simple transfer learning technique from an existing large-scale Amazon community question answering dataset. Transfer learning can result in an increase of about 10\% in accuracy in the experimental setting where we restrict the size of the data of the target task used for training. As an application of this work, we integrate the best performing model trained in this work into a mobile-based shopping assistant and
show its usefulness.
\end{abstract}


\section{Introduction}
\begin{figure*}
  \centering
  \includegraphics[width=\linewidth,keepaspectratio,frame]{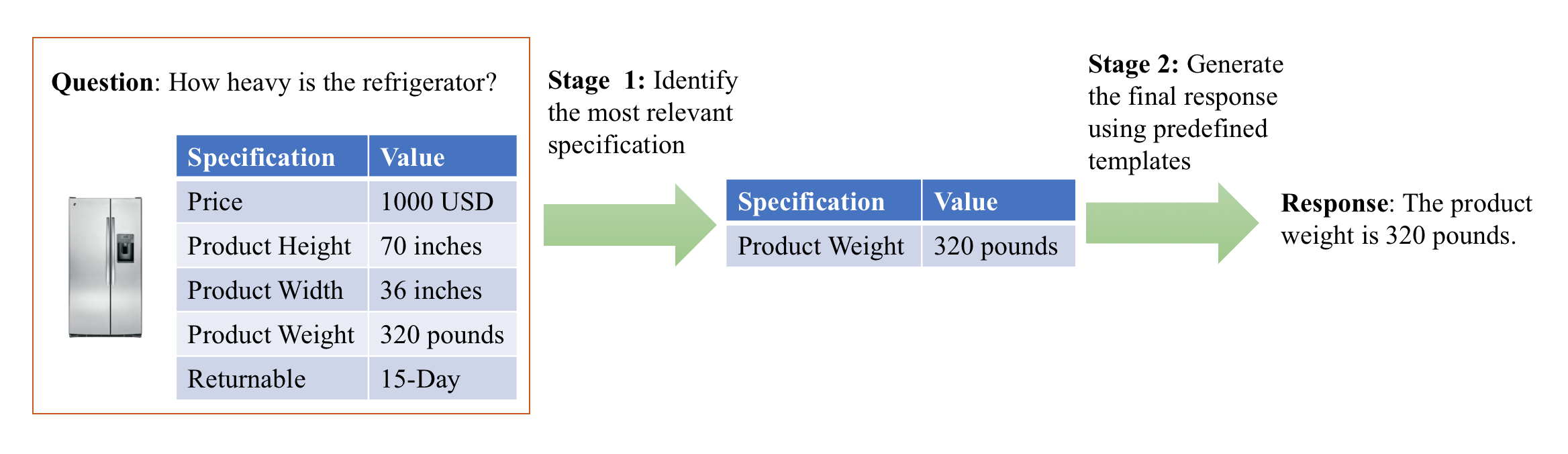}
  \caption{An approach to answering questions related to product facts and specifications. In this work, we focus on the first stage of the approach, which is similar to the answer selection problem.}
  \label{fig:overall_process}
\end{figure*}

Customers ask many questions before buying products. They want to get adequate information to determine whether the product of interest is worth their money. Because the questions customers ask are diverse, developing a general question answering system to assist customers is challenging. In this paper, we are particularly interested in the task of answering questions regarding product facts and specifications. We formalize the task as follows: Given a question \(Q\) about a product \(P\) and the list of specifications \((s_{1}, s_{2}, ..., s_{M})\) of \(P\), the goal is to identify the specification that is most relevant to \(Q\). \(M\) is the number of specifications of \(P\), and \(s_{i}\) is the \(i^{th}\) specification of \(P\). In this formulation, the task is similar to the answer selection problem \cite{Wang2017BilateralMM,Bian2017ACM,Shen2017InterWeightedAN,Tran2018TheCA,C18-1181}. `Answers' shall be individual product specifications in this case. In production, given a question about a product, a possible approach is to first select the specification of the product that is most relevant to the question and then use the selected specification to generate the complete response sentence using predefined templates. Figure \ref{fig:overall_process} illustrates the approach.

Many e-commerce websites offer community question answering (CQA) systems for users to pose product-related questions and experienced customers may provide answers voluntarily. In popular websites such as Amazon or eBay, the amount of CQA data (i.e., questions and answers) is huge and growing over time. For example, in \cite{Wan2016ModelingAS}, the authors could collect a QA dataset consisting of 800 thousand questions and over 3.1 million answers from the CQA platform of Amazon.

In this paper, we show that the large amount of CQA data of popular e-commerce websites can be used to improve the performance of models for answering questions related to product facts and specifications. Our experimental results demonstrate that the performance of a model for answering questions related to products listed in the Home Depot website can be improved by a large margin via a simple transfer learning technique from an existing large-scale Amazon community question answering dataset. Transfer learning can result in an increase of about 10\% in accuracy in the experimental setting where we restrict the size of the data of the target task used for training. As an application of this work, we integrate the best performing model trained in this work into a mobile-based shopping assistant and
show its usefulness.

\section{Related Work}

\subsection{Recurrent Neural Networks}
Recurrent Neural Networks (RNNs) form an expressive model family for processing sequential data. They have been widely used in many tasks, including machine translation \cite{Cho2014LearningPR,Luong2015EffectiveAT}, image captioning \cite{Karpathy2015DeepVA}, and document classification \cite{Yang2016HierarchicalAN}. The Long Short Term Memory (LSTM) \cite{Hochreiter1997LongSM} is one of the most popular variations of RNN. The main components of the LSTM are three gates:  an input gate $i_{t}$ to regulate the information flow from the input to the memory cell, a forget gate $f_{t}$ to regulate the information flow from the previous time step's memory cell, and an output gate that regulates how the model produces the outputs from the memory cell. Given an input sequence $\{\boldsymbol{x}_{1}, \boldsymbol{x}_{2}, ..., \boldsymbol{x}_{n}\}$ where ${\boldsymbol{x}_{t}}$ is typically a word embedding, the computations of LSTM are as follows:

\begin{equation}
\begin{aligned}
i_{t} &= \sigma(\boldsymbol{W}_{i}\boldsymbol{x}_{t} + \boldsymbol{U}_{i}\boldsymbol{h}_{t-1} + \boldsymbol{b}_{i})\\
f_{t} &= \sigma(\boldsymbol{W}_{f}\boldsymbol{x}_{t} + \boldsymbol{U}_{f}\boldsymbol{h}_{t-1} + \boldsymbol{b}_{f})\\
o_{t} &= \sigma(\boldsymbol{W}_{o}\boldsymbol{x}_{t} + \boldsymbol{U}_{o}\boldsymbol{h}_{t-1} + \boldsymbol{b}_{o})\\
\widetilde{C}_{t} &= tanh(\boldsymbol{W}_{c}\boldsymbol{x}_{t} + \boldsymbol{U}_{c}\boldsymbol{h}_{t-1} + \boldsymbol{b}_{c})\\
{C}_{t} &= i_{t} \odot \widetilde{C}_{t} + f_{t} \odot {C}_{t-1}\\
\boldsymbol{h}_{t} &= o_{t} \odot {C}_{t}
\end{aligned}
\end{equation}

The standard single direction LSTM
processes input only in one direction, it does not
utilize the contextual information from the future inputs. In other words, the value of $\boldsymbol{h}_{t}$ does not depend on any element of $\{\boldsymbol{x}_{t+1}, \boldsymbol{x}_{t+2}, ..., \boldsymbol{x}_{n}\}$. On the other hand, a bi-directional LSTM (biLSTM) utilizes both the previous and future context by processing the input sequence on two directions, and generate two independent sequences of LSTM output vectors. One processes the input sequence in the forward direction, while the other processes the input in the reverse direction. The output at each time step is the concatenation of the two output vectors from both directions, i.e., $\boldsymbol{h}_{t} = \vec{\boldsymbol{h}_{t}}\,||\,\cev{\boldsymbol{h}_{t}}$. In this case, the value of $\boldsymbol{h}_{t}$ will depend on every element of $\{\boldsymbol{x}_{1}, \boldsymbol{x}_{2}, ..., \boldsymbol{x}_{n}\}$.

\subsection{Answer Selection}
Answer selection is an important problem in natural language processing. Given a question and a set of candidate answers, the task is to identify which of the candidates contains the correct answer to the question. For example, given the question ``Who won the Nobel Prize in Physics in 2016?" and the following candidate answers:
\begin{enumerate}
  \item The Fields Medal is regarded as one of the highest honors a mathematician can receive, and has been described as the mathematician's Nobel Prize.
  \item Neutron scattering played an important role in the experimental exploration of the theoretical ideas of Thouless, Haldane, and Kosterlitz, who won the Nobel Prize in Physics in 2016.
  \item The Nobel Prize was established in the will of Alfred Nobel, a Swedish inventor of many inventions, most famously dynamite.
\end{enumerate}
The second answer should be selected. 

Previous work on answer selection typically relies on feature engineering, linguistic tools, or external resources \cite{trec_qa_dataset,Wang:2010:PTM:1873781.1873912,Heilman:2010:TEM:1857999.1858143,question-answering-using-enhanced-lexical-semantic-models,Yao13answerextraction}. Recently, many researchers have investigated employing deep learning for the task \cite{Wang2017BilateralMM,Bian2017ACM,Shen2017InterWeightedAN,Tran2018TheCA}. Most recently proposed deep learning models outperform traditional techniques. In addition, they do not need any feature-engineering effort or hand-coded resources beyond some large unlabeled corpus on which to learn the initial word embeddings, such as word2vec \cite{Mikolov_word2vec} or GloVe \cite{Pennington2014GloveGV}. The authors in \cite{C18-1181} provide a comprehensive review on deep learning methods applied to answer selection. In addition, the most popular datasets and the evaluation metrics for answer selection are also described in the work.

\subsection{Customer Service Chatbots}
Developing customer service chatbots for ecommerce websites is an active area. For example, ShopBot \footnote{https://shopbot.ebay.com} aims at helping consumers narrow down the best deals from eBay's over a billion listings. SuperAgent introduced in \cite{SuperAgentMicrosoft} is a powerful chatbot designed to improve online
shopping experience. Given a specific product page and a customer question, SuperAgent selects the best answer from multiple data sources within the page such as in-page product information, existing customer questions \& answers,
and customer reviews of the product. In \cite{W18-3105}, T. Lai et al. proposed a simple but effective deep learning model for answering questions regarding product facts and specifications.

\subsection{Transfer Learning for Question Answering}
Transfer learning \cite{5288526} has been successfully applied to various domains such as speech recognition \cite{6639081}, computer vision \cite{Razavian:2014:CFO:2679599.2679731}, and natural language processing \cite{DBLP:journals/corr/ZhangBJ17}. Its applicability to question answering and answer selection has recently been studied \cite{Min2017QuestionAT,DBLP:journals/corr/abs-1711-05345}. In \cite{Min2017QuestionAT}, the authors created SQuAD-T, a modification of the original large-scale SQuAD dataset \cite{Rajpurkar2016SQuAD10} to allow for directly training and evaluating answer selection systems. Through a basic transfer learning technique from SQuAD-T, the authors achieve the state of the art in two well-studied QA datasets, WikiQA \cite{Yang2015WikiQAAC} and SemEval-2016 (Task 3A) \cite{Nakov2016SemEval2016T3}. In \cite{DBLP:journals/corr/abs-1711-05345}, the authors tackle the TOEFL listening comprehension test \cite{Tseng2016TowardsMC} and MCTest \cite{Richardson2013MCTestAC} with transfer learning from MovieQA \cite{Tapaswi2016MovieQAUS} using two existing QA models. To the best of our knowledge, there is no published work on exploring transfer learning techniques for improving the performance of models for answering questions related to product facts and specifications.

\section{Approach}
\subsection{Baseline Model}
\begin{figure*}
  \centering
  \includegraphics[width=0.8\linewidth,keepaspectratio,frame]{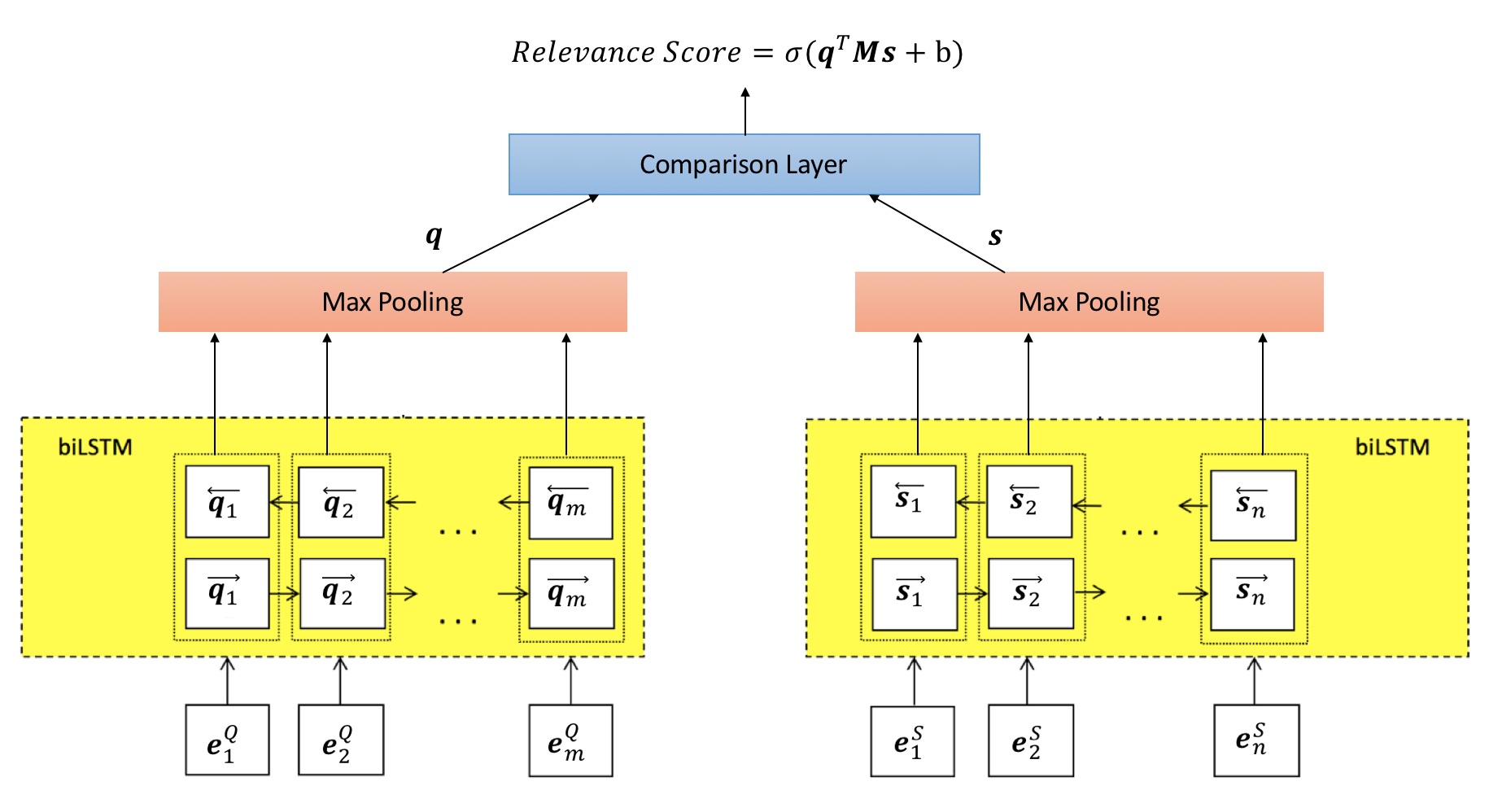}
  \caption{Architecture of the baseline model.}
  \label{fig:baseline_model}
\end{figure*}

Our task of matching questions and product specifications is similar to the answer selection problem. ``Answers'' shall be individual product
specifications. Even though a common trait of a number of recent state-of-the-art methods for answer selection is the use of complicated deep learning models \cite{Wang2017BilateralMM,Bian2017ACM,Shen2017InterWeightedAN,Tran2018TheCA}, T. Lai et al. showed in \cite{W18-3105} that complicated models may not be needed in this case. Simple but well designed models may match the performance of complicated models for the task of selecting the most relevant specification to a question. Inspired by the results of \cite{W18-3105}, we propose a new simple baseline model for the task.

The baseline model takes a question and a specification name as input and outputs a score indicating their relevance. During inference, given a question, the model is used to assign a score to every candidate specification based on how relevant the specification is. After that, the top-ranked specification is selected. Figure \ref{fig:baseline_model} illustrates the overall architecture of the baseline model. Given a question $Q$ and a specification name $S$, the model first transforms $Q$ and $S$ into two sequences \(Q_{e} = [\textbf{e}_{1}^{Q}, \textbf{e}_{2}^{Q}, ..., \textbf{e}_{m}^{Q}]\) and \(S_{e} = [\textbf{e}_{1}^{S}, \textbf{e}_{2}^{S}, ..., \textbf{e}_{n}^{S}]\) using word embeddings pre-trained with GloVe \cite{Pennington2014GloveGV}. Here, \(\textbf{e}_{i}^{Q}\) is the embedding of the \(i^{th}\) word of the question and \(\textbf{e}_{j}^{S}\) is the embedding of the \(j^{th}\) word of the specification name. \(m\) and \(n\) are the lengths of \(Q\) and \(S\), respectively. After that, we feed $Q_{e}$ and $S_{e}$ individually into a parameter shared biLSTM model. For the question $Q$, we will obtain a sequence of vectors $[\textbf{q}_{1},\textbf{q}_{2},...,\textbf{q}_{m}]$ from the biLSTM where $\textbf{q}_{i} = \vec{\textbf{q}_{i}}\;||\;\cev{\textbf{q}_{i}}$. To form a final fixed-size vector representation of the question $Q$, we select the maximum value over
each dimension of the vectors $[\textbf{q}_{1},\textbf{q}_{2},...,\textbf{q}_{m}]$ (max pooling). We denote the final representation of the question as $\textbf{q}$. In a similar way, we can obtain the final representation $\textbf{s}$ of the specification. Finally, the probability of the specification being relevant is

\begin{equation}
p(y = 1|\textbf{q}, \textbf{s}) = \sigma(\textbf{q}^{T}\;\textbf{M}\;\textbf{s} + b)
\end{equation}
where the bias term $b$ and the transformation matrix $\textbf{M}$ are model parameters. The sigmoid function squashes the score to a probability between 0 and 1.

\subsection{Transfer Learning}
The transfer learning technique used in this work is simple and includes two steps. The first step is to pre-train the baseline model on a large \textbf{source dataset}. The second step is to fine-tune the same model on the \textbf{target dataset}, which typically contains much less data than the source dataset. The effectiveness of transfer learning is evaluated by the performance of the baseline model on the target dataset.

\subsection{Datasets}
\begin{table*}[!ht]
\centering
\small
\begin{tabular}{|c|c|c|c|}
\hline Product ID & Product Category & Question & Correct Specification\\ \hline

207025690 & Microwaves  & What is the wattage? & Wattage (watts) \\\hline

205209621 & Refrigerators & How many bottles can I place inside? & Bottle Capacity\\\hline

301688014 & Smart TVs & Is the screen size at least 50 inches? & Screen Size (In.) \\\hline

205867752 & Electric Ranges & Can I return the range if I change my mind? & Returnable\\\hline

\end{tabular}
\caption{\label{homedepotqa_examples} Examples of correct question-specification pairs.}
\end{table*}

\begin{table*}[!ht]
\centering
\small
\begin{tabular}{|c|c|c|c|}
\hline Split & \# Positive Pairs & \# Negative Pairs & \# Pairs \\ \hline

Training Set &  815484 & 812030 & 1627514 \\\hline

Development Set & 2482 & 2518 & 5000 \\\hline

\end{tabular}
\caption{\label{amazoncqa_statistics} Statistics of the AmazonCQA dataset.}
\end{table*}

\textbf{HomeDepotQA.} The target dataset used for experiments is created using
Amazon Mechanical Turk (MTurk) \footnote{https://www.mturk.com}. MTurk connects requesters (i.e., people who have works to be done) and workers (i.e., people who work on tasks for money). Requesters can post small tasks for workers to complete for a fee. These small tasks are referred to as human intelligence tasks (HITs). We crawled the information of products listed in the Home Depot website \footnote{https://www.homedepot.com}. For each product, we create HITs where workers are asked to write questions regarding the specifications of the product. The final dataset consists of 7,119 correct question-specification pairs that
are related to 153 different products in total. Table \ref{homedepotqa_examples} shows some examples of correct question-specification pairs collected. We refer to the dataset as \textbf{HomeDepotQA}. We split up the dataset into training
set, development set, and test set so that the
test set has no products in common with the training
set or the development set. For example, if the test set has a question about the product with ID 205148717, then there will be no questions about that product in the training set or the development set. We are interested in whether our proposed model can be generalized to answer questions about new products. The proportions
of the training set, development set, and
test set are roughly 80\%, 10\%, and 10\% of the total correct question-specification pairs, respectively.

\textbf{AmazonCQA.} The source dataset used for experiments is a preprocessed version of the QA dataset collected in \cite{Wan2016ModelingAS}. The original dataset consists of 800 thousand questions and over 3.1 million answers collected from the CQA platform of Amazon. Each answer is supposed to address its corresponding question. But since the CQA data (i.e., questions and answers) is a resource created by a community of casual users, there is a lot of noise in addition to the complications of informal language use, typos, and grammatical mistakes. Below are the major preprocessing steps applied to the original dataset:
\begin{enumerate}
\item We remove questions or answers that contain URLs, because the target dataset does not have any questions or specification names that contain URLs.
\item We set the minimum length of questions to four tokens for filtering out poorly structured questions. There can be many examples where the questions are very short and not grammatically correct; for example, people might just ask: ``Waterproof?''. In the target dataset, a question is typically a complete sentence (e.g., ``What happens if this laser level kit gets wet?'').
\item Answers must also contain at least ten tokens, as the same problem can occur here; for example, the answer might be a single ``Yes", which does not contain much semantic information.
\item We remove questions or answers that are too long. One reason is that most of the specification names in the target dataset are short. Therefore, the answers in the source dataset should not be too long.
\item We remove answers that contain phrases such as ``I have no idea'' or ``I am not sure'', because it is likely that those answers do not contain any information relevant to the question.
\item In order to be able to train the baseline model, we sample one or more negative answers for each question. 
\end{enumerate}
We refer to the final preprocessed dataset as \textbf{AmazonCQA}. We split up the dataset into a training set and a development set. We do not need a test set as the effectiveness of transfer learning is evaluated by only the performance on the target dataset. The statistics of the AmazonCQA dataset is shown in Table \ref{amazoncqa_statistics}.

\section{Experiments and Results}
\begin{table*}[!ht]
\centering
\small
\begin{tabular}{|c|c|c|c|}
\hline Pre-trained & Percentage of HomeDepotQA's train set used & MRR & Accuracy \\ \hline

No & 10\% & 0.7636 & 0.6667 \\ \hline
Yes & 10\% & \textbf{0.8442} & \textbf{0.7656}  \\ \hline
\hline
No & 50\% & 0.8636 & 0.7933  \\ \hline
Yes & 50\% & \textbf{0.9049} & \textbf{0.8486} \\ \hline
\hline
No & 100\% & 0.8815 & 0.8180  \\ \hline
Yes & 100\% & \textbf{0.9030} & \textbf{0.8443} \\\hline

\end{tabular}
\caption{\label{results} Results of transfer learning on the target datasets.}
\end{table*}

During the pre-training step, we trained the baseline model on the training set of AmazonCQA. We chose the hyperparameters of the model using the development set of  AmazonCQA. After that, during fine-tuning, the model was further trained on the training set of the target dataset (i.e., HomeDepotQA) and tuned on the development set, and the performance on the testing set of the target dataset was reported as the final result. We use mean reciprocal rank (MRR) and  accuracy as the performance measurement metrics. In addition to fine-tuning the baseline model on the entire training set of HomeDepotQA, we conducted experiments where we restricted the amount of training data from HomeDepotQA for fine-tuning the model. Table \ref{results} shows the experiment results. Pre-training on the AmazonCQA dataset clearly helps. In the setting where only 10\% of the correct question-specification pairs in the training set of HomeDeptQA is used, transfer learning can result in an increase of about 10\% in accuracy.

It is worth mentioning that when training on the train set of HomeDepotQA, we use all possible question-specification pairs. In other words, if there are $k$ questions about a product and the product has $h$ specifications, then there are $h \times k$ question-specification examples related to the product, and exactly $k$ of them are positive examples.

\section{Application}
As an application of this work, we integrate the best performing model trained in this work into ISA, a mobile-based intelligent shopping
assistant. ISA is designed to improve users' shopping experience in brick and mortar stores. First, an in-store user just needs to take a picture or scan the barcode of a product. ISA then retrieves the information of the product of interest from a database
by using computer vision techniques. After that,
the user can ask natural language questions about the product to ISA. The user can either type in the
questions or directly speak out the questions using voice. ISA is integrated with both speech recognition and speech synthesis abilities, which allows users to ask questions without typing.

\begin{figure}[!htb]
  \centering
  \includegraphics[width=\linewidth,height=10cm,keepaspectratio,frame]{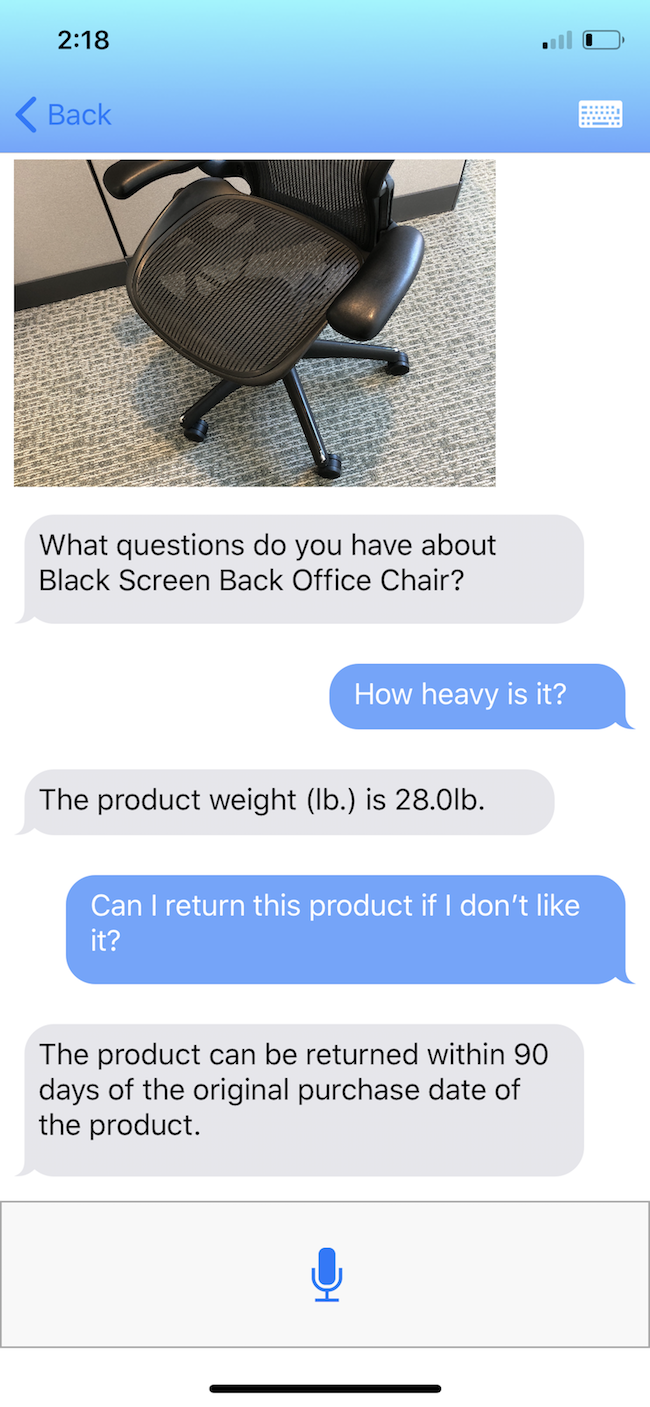}
  \caption{Answering questions regarding product specifications.}
  \label{fig:product_qa_output}
\end{figure}

The role of our model is to answer questions regarding the specifications of a product. Given
a question about a product, the model is used to
rank every specification of the product based on
how relevant the specification is. We select the
top-ranked specification and use it to generate the
response sentence using predefined templates. Figure \ref{fig:product_qa_output} shows example outputs generated
using our model.

\section{Conclusion}
In this work, we show that the large volume of existing CQA data can be beneficial when building a system for answering questions related to product facts and specifications. Our experimental results demonstrate that the performance of a model for answering questions related to products listed in the Home Depot website can be improved by a large margin via a simple transfer learning technique from an existing large-scale Amazon CQA dataset. Transfer learning can result in an increase of about 10\% in accuracy in the experimental setting where we restrict the size of the data of the target task used for training. In addition, we also integrate the best performing model trained in this work into ISA, an intelligent shopping assistant that is designed with the goal of improving shopping experience in physical stores. In the future, we plan to investigate more transfer learning techniques for utilizing the large volume of existing CQA data. 

\bibliographystyle{IEEEtran}
\bibliography{IEEEabrv,mybibfile}

\end{document}